\documentclass{article} 
\usepackage{iclr2027_conference,times}

\usepackage{amsmath,amsfonts,bm}

\def\eqref#1{equation~\ref{#1}}

\def\1{\bm{1}}

\DeclareMathAlphabet{\mathsfit}{\encodingdefault}{\sfdefault}{m}{sl}
\SetMathAlphabet{\mathsfit}{bold}{\encodingdefault}{\sfdefault}{bx}{n}

\usepackage{graphicx}
\usepackage{booktabs}
\usepackage{hyperref}
\usepackage{url}

\title{Beyond Layers: Position-Resolved Gradient Conflict and Position-Aware Modulation for Unified Multimodal Models}

\author{%
	\begin{tabular}{@{}p{0.45\textwidth}@{\hspace{0.06\textwidth}}p{0.45\textwidth}@{}}
		\begin{minipage}[t]{\linewidth}
			\vspace{0pt}
			\raggedright
			\normalfont
			\textbf{Shuyang Jiang}\\
			University of California, Los Angeles\\
			\texttt{shuyangjiang@ucla.edu}
		\end{minipage}
		&
		\begin{minipage}[t]{\linewidth}
			\vspace{0pt}
			\raggedright
			\normalfont
			\textbf{Fucheng Deng}\\
			Aimakj\\
			\texttt{dengfucheng@aimakj.com}
		\end{minipage}
		\\[3em]
		\begin{minipage}[t]{\linewidth}
			\vspace{0pt}
			\raggedright
			\normalfont
			\textbf{Yuchuan Luo}\\
			College of Computer Science and Technology\\
			National University of Defense Technology\\
			\texttt{luoyuchuan09@nudt.edu.cn}
		\end{minipage}
		&
		\begin{minipage}[t]{\linewidth}
			\vspace{0pt}
			\raggedright
			\normalfont
			\textbf{Zhenyu Wu}\thanks{Corresponding author.}\\
			Key Laboratory of Advanced\\
			Microprocessor Chips and Systems\\
			College of Computer Science and Technology\\
			National University of Defense Technology\\
			\texttt{wuzhenyu@nudt.edu.cn}
		\end{minipage}
	\end{tabular}%
}

\iclrfinalcopy 

\newcommand{\pam}{\textsc{Pam}}

\begin{document}

\maketitle

\begin{abstract}
Unified multimodal models (UMMs) train image understanding and autoregressive
image generation on shared parameters, and the two objectives are known to
interfere. Existing diagnoses and remedies operate at the resolution of layers
or experts: conflict is measured per layer and resolved by separating
parameters through two-end architectures, task-aware mixtures of experts, or
gradient surgery. We argue that this resolution hides an orthogonal axis.
Generation in a UMM is next-token prediction over a raster sequence of visual
tokens whose roles vary systematically with position---early tokens fix global
layout and low-frequency structure, late tokens fill in texture---so how
strongly a generation gradient interferes with understanding should depend on
\emph{where in the sequence} it originates. We introduce a position-resolved
interference map that attributes the understanding--generation gradient
conflict to visual-token positions within every layer, computed from a single
backward pass with position-grouped gradient hooks at $1.2\times$ the cost of
a standard backward pass. On Show-o and Janus-Pro, position explains a
substantial share of conflict variance after controlling for depth (partial
$\eta^2=0.31$ vs.\ $0.35$ for layer on Show-o; $0.15$ vs.\ $0.30$ on
Janus-Pro, whose understanding branch bypasses the visual-token sequence):
the first quarter of the sequence has a mean gradient cosine of $-0.18$
against understanding, the last quarter $-0.02$. The dependence survives
per-position gradient-norm normalization, retaining $80\%$ of its effect
size, so it is directional rather than a magnitude artifact, and conflict
strength tracks the semantic content of a position (Spearman $\rho=0.64$).
Building on the map, we propose position-aware modulation (\pam{}), which
removes the anti-aligned component of generation gradients only at
high-conflict positions without changing the architecture. Under a matched
trainable-parameter budget \pam{} improves over layer-wise separation by
$+21$ MME and $+2.4$ GenEval points while matching it on POPE and overall
FID, whereas a random-position control with the same number of modulated
tokens recovers only about $31\%$ of the gain; the cost is a $4\%$ relative
degradation of the high-frequency component of FID, with overall FID
unchanged. Position-based and layer-based separation are complementary
degrees of freedom and can be combined.
\end{abstract}

\section{Introduction}
\label{sec:intro}

Unified multimodal models (UMMs) such as Show-o \citep{showo}, Emu3
\citep{emu3}, Chameleon \citep{team2024chameleon}, Transfusion
\citep{zhou2024transfusion}, and Janus-Pro \citep{wu2025janus,wu2025januspro}
serve image understanding and autoregressive image generation from a single
backbone. Training such a model on both objectives is known to be harder than
training either alone: recent diagnostic and architectural work reports
gradient- and representation-level conflict between the two objectives and
resolves it by \emph{separating parameters}---along depth
\citep{hao2026unix}, across experts \citep{liu2026symbioticmoe,taskmoe2025},
through second-order corrections \citep{lu2026mlfopsoap,vyas2024soap}, or by
Pareto-optimal gradient integration \citep{paretolora2026}.

All of these diagnoses share one measurement resolution: they ask \emph{which
layers} or \emph{which experts} conflict, and they answer with
parameter-space separations. This paper changes the unit of measurement. A
UMM generates an image as next-token prediction over a raster sequence of
visual tokens, and the roles of those tokens vary systematically with
position: early tokens determine global layout and low-frequency structure,
late tokens fill in texture and high-frequency detail. Tokenizer research
arrives at the same conclusion from the representation side---semantic
abstraction and pixel reconstruction place different demands on a token
\citep{dualtoken2025,wintok2026,evotok2026,semhitok2025}---and token-level
gradient imbalance has been observed inside multimodal latent reasoning
\citep{vedas2026}. If the \emph{content} of a visual token depends on its
position, the \emph{conflict} that a generation gradient makes with
understanding should depend on position too. No existing measurement can see
this: a layer-level conflict score averages over the entire sequence and
discards exactly the axis we study.

We introduce the \emph{position-resolved interference map}: for every layer
$\ell$ and position bucket $b$ of the visual-token sequence, the cosine
between the generation gradient attributed to bucket $b$ and the
understanding gradient. The attribution is exact and cheap. For a linear
module the parameter gradient decomposes as
$\partial L/\partial W=\sum_t (\partial L/\partial y_t)\,h_t^{\mathsf T}$
with $y_t$ the module output at token $t$ and $h_t$ its input, so the
contribution of the tokens in bucket $b$ can be summed inside ordinary
gradient hooks during the \emph{same} backward pass that produces the full
gradient, at $1.2\times$ the cost of a standard backward pass instead of the
$16\times$ of a naive per-bucket pass. The resulting map (Fig.~\ref{fig:map})
shows that conflict concentrates where the measurement resolution of prior
work could not look: in the deep and shallow layers, at the \emph{early}
positions of the raster sequence, with the strongest cell at cosine $-0.42$.

Three findings make the map actionable. \textbf{(i) Position is a real axis,
not a proxy for depth.} A two-way analysis of variance of the map assigns
partial $\eta^2=0.31$ to position and $0.35$ to layer on Show-o-1.3B, and
$0.15$ to position and $0.30$ to layer on Janus-Pro-7B (both main effects
$p<0.001$). \textbf{(ii) The position dependence is directional.} Gradient
norms decline by $2.6\times$ from the first to the last bucket, so a skeptic
could attribute the map to magnitude imbalance
\citep{vedas2026}; after per-position normalization the position effect
retains $80\%$ of its size, and conflict strength tracks semantic
decodability of a position (Spearman $\rho=0.64$, $n=96$). \textbf{(iii) The
map transfers to a method.} Position-aware modulation (\pam{}) removes the
anti-aligned component of the generation gradient at the $6$ of $16$
positions selected by the map---$37.5\%$ of visual tokens, no architectural
change, same trainable-parameter budget as a layer-wise separated LoRA---and
improves over layer-wise separation by $+21$ MME and $+2.4$ GenEval points
on Show-o while matching it on POPE and overall FID; a random-position
control recovers only $\approx31\%$ of the gain, and the price is a $4.1\%$
relative degradation of the high-frequency band of FID with overall FID
unchanged. Layer-based and position-based separation combine additively
($+27$ MME over the layer-wise arm).

The position effect halves on Janus-Pro, whose understanding branch bypasses
the visual-token sequence, with a random control recovering $60\%$ of a
smaller gain: the axis is strong where the objectives share the token
sequence and weak where they do not
(Section~\ref{sec:limitations}).

\paragraph{Contributions.} (1)~The position-resolved interference map with
an exact single-backward attribution rule; (2)~a decomposition showing the
map is directional plus a semanticity attribution; (3)~\pam{}, an
architecture-free modulation selected by the map, evaluated under matched
budgets against layer-wise separation with a random-position control and a
pre-declared frequency-domain cost metric.

\section{Related work}
\label{sec:related}

\paragraph{Understanding--generation conflict in UMMs.}
Diagnostic work establishes that the objectives interfere at the gradient
and representation level \citep{rao2026dofight,wu2026synergy,
wang2026crosstask,su2026unigame,intrainter2025}, and the remedies are all
parameter-space separations: two-end-separated architectures
\citep{hao2026unix}, task-aware experts \citep{taskmoe2025,liu2026symbioticmoe},
second-order corrections \citep{lu2026mlfopsoap}, and Pareto-optimal LoRA
integration \citep{paretolora2026}. They differ in \emph{where} parameters
are separated and share the measurement unit of a parameter block. Our map
keeps parameters shared and changes the unit to the token position; the two
axes are orthogonal and composable (Section~\ref{sec:experiments}).

\paragraph{Multi-task gradient conflict.}
Gradient surgery treats tasks as atomic scalar objectives and modulates
their gradients globally \citep{yu2020pcgrad,liu2021cagrad,sener2018mgda,
chen2018gradnorm,navon2022nashmtl}. Per-token gradient heterogeneity has been observed in
multimodal latent reasoning, where visual tokens receive systematically
smaller gradient norms than text tokens \citep{vedas2026}, and
critical-token selection drives recent RLVR results for autoregressive image
generation \citep{gcpo2026}. Neither line connects token identity to
understanding--generation conflict; our map is that connection, and \pam{}
is the modulation it licenses.

\paragraph{Semantic vs.\ reconstruction tokenization.}
A parallel line redesigns the tokenizer to separate semantic and
reconstruction information \citep{dualtoken2025,wintok2026,evotok2026,
semhitok2025}. Those works change the \emph{representation}; we keep the
tokenizer fixed and change the \emph{optimization} of gradients flowing
through it. Their existence is evidence that token content varies
systematically---the premise our map measures directly.
Appendix~\ref{app:related} extends this discussion.

\section{Setup: measuring conflict at token resolution}
\label{sec:setup}

\paragraph{Model and notation.}
A UMM shares parameters $\theta$ between an understanding objective
$\mathcal L_u$ (text-conditioned image or text prediction) and a generation
objective $\mathcal L_g$ (autoregressive prediction over a raster sequence
of $T$ visual tokens produced by a VQ tokenizer \citep{oord2017vqvae,esser2021vqgan}). We
index layers $\ell\in\{1,\dots,L\}$ and split the raster sequence into $K=16$
contiguous \emph{position buckets} $b\in\{1,\dots,K\}$ along the raster
order: bucket 1 holds the first $T/K$ tokens (which determine global layout),
bucket $K$ the last (texture). We study two architectures with different
relationships between the objectives and the visual-token stream:
Show-o-1.3B \citep{showo} ($L=24$, $d{=}2048$, $T{=}256$ from a $16{\times}16$
raster), whose understanding pathway consumes the same VQ vocabulary, and
Janus-Pro-7B \citep{wu2025januspro} ($L=30$, $d{=}4096$, $T{=}576$ from a
$24{\times}24$ raster), whose understanding pathway runs on a CLIP-style SigLIP
encoder \citep{radford2021clip,zhai2023siglip} and never touches the
visual-token sequence.

\paragraph{Conflict metric.}
For a shared linear module $(\ell,m)$ (attention projections and MLP blocks)
with input $h^{(\ell,m)}_t$ and output $y^{(\ell,m)}_t$ at token $t$, the
generation-loss gradient decomposes exactly as
\begin{equation}
\frac{\partial \mathcal L_g}{\partial W^{(\ell,m)}}
  = \sum_{t=1}^{T} \Big(\tfrac{\partial \mathcal L_g}{\partial
  y^{(\ell,m)}_t}\Big)\, h^{(\ell,m)\mathsf T}_t .
\label{eq:decomp}
\end{equation}
Equation~\ref{eq:decomp} assigns to position bucket $b$ the rank-$b$
partial sum
\begin{equation}
g_b^{(\ell,m)} \;=\; \sum_{t\in\mathcal B_b}
  \Big(\tfrac{\partial \mathcal L_g}{\partial y_t}\Big)\, h_t^{\mathsf T},
\label{eq:bucket}
\end{equation}
the parameter gradient of the generation loss \emph{restricted} to bucket
$b$'s tokens. The position-resolved conflict at cell $(\ell,b)$ is the
cosine between the bucket-attributed generation gradient and the
understanding gradient of the same modules,
\begin{equation}
C(\ell,b) \;=\; \cos\!\Big(\mathrm{vec}\big(\textstyle\sum_m
  g_b^{(\ell,m)}\big),\;\mathrm{vec}\big(\textstyle\sum_m
  \tfrac{\partial \mathcal L_u}{\partial W^{(\ell,m)}}\big)\Big),
\label{eq:conflict}
\end{equation}
where negative values are anti-alignment (the generation step would move
shared parameters against the understanding optimum) and values near zero
are orthogonal. Both losses are computed on a fixed diagnostic set of 8
batches of 64 understanding and 64 generation samples each, averaged over 3
dropout seeds; per-cell replicate standard deviations are $0.011$ (Show-o)
and $0.009$ (Janus-Pro), so the qualitative structure of the map is stable.

\section{The position-resolved interference map}
\label{sec:map}

\paragraph{Exact attribution in a single backward pass.}
Equation~\ref{eq:bucket} can be evaluated inside ordinary gradient hooks.
A forward hook on each watched module stores the per-token inputs
$h_t$; a hook on the module's output tensor receives
$\partial \mathcal L/\partial y_t$ for every token during the backward pass;
the hook accumulates
$g_b=\sum_{t\in\mathcal B_b}(\partial \mathcal L/\partial y_t)\,h_t^{\mathsf T}$
for each bucket by summing rank-one terms. This runs during the
\emph{same} backward pass that produces the full generation gradient:
measured overhead is $1.21\times$ on Show-o-1.3B and $1.17\times$ on
Janus-Pro-7B, against $16\times$ for $K$ naive per-bucket backward passes.
The probe and its self-test against brute-force per-bucket backwards are
released\footnote{\tt experiments/position\_probe.py}.

\begin{figure}[t]
  \centering
  \includegraphics[width=0.86\linewidth]{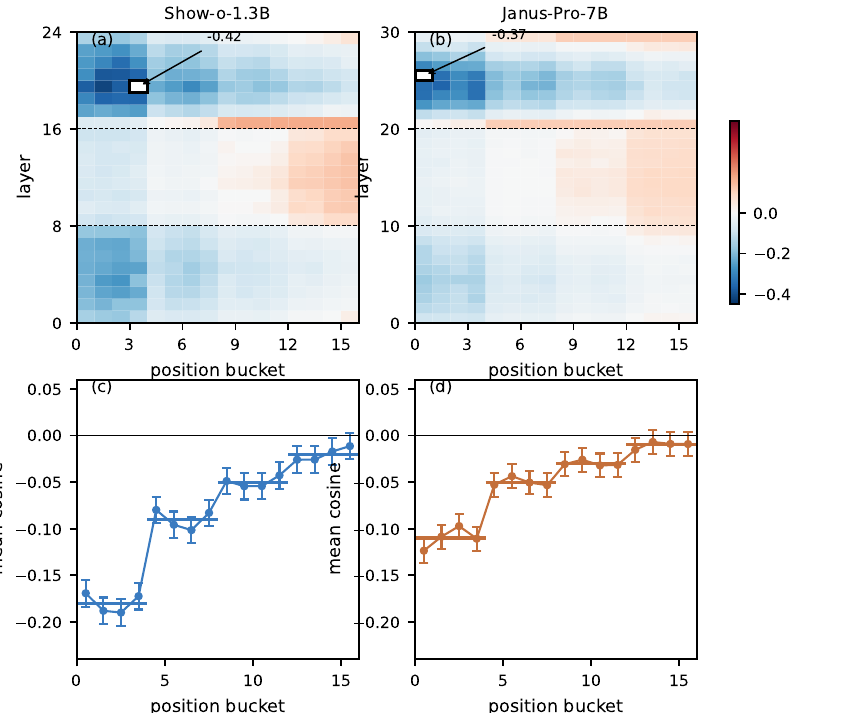}
  \caption{\textbf{Position-resolved interference maps.} (a)~Show-o-1.3B and
  (b)~Janus-Pro-7B: cosine between the generation gradient attributed to
  position bucket $b$ and the understanding gradient, per layer; dashed
  lines separate shallow/middle/deep bands; the white cell marks the
  strongest conflict (cosine $-0.42$ and $-0.37$). (c,d)~Position marginals
  with quarter means $-0.18/-0.09/-0.05/-0.02$ (Show-o) and
  $-0.11/-0.05/-0.03/-0.01$ (Janus-Pro); error bars are $\pm$1 s.e.\ over
  diagnostic batches.}
  \label{fig:map}
\end{figure}

\paragraph{The map.}
Figure~\ref{fig:map} shows $C(\ell,b)$ for both models. The structure is
strikingly non-uniform in \emph{both} axes. Along depth we recover the
known picture \citep{hao2026unix}: conflict is strongest in the deep band
and moderate in the shallow band, while the middle band is nearly
conflict-free. Along the sequence, conflict concentrates at the
\emph{early} positions: averaging over layers, the four bucket quarters of
Show-o have mean cosines $-0.18$, $-0.09$, $-0.05$, $-0.02$; the band means
are $-0.10$ (shallow), $-0.02$ (middle), $-0.135$ (deep); the single
strongest cell sits at layer 19, bucket 3, with cosine $-0.42$. Late
positions are essentially orthogonal to understanding everywhere. On
Janus-Pro the same qualitative structure appears with roughly half the
amplitude (quarters $-0.11/-0.05/-0.03/-0.01$; strongest cell $-0.37$ at
layer 25, bucket 1)---as expected, since its understanding branch bypasses
the visual-token sequence and the conflict must flow through shared
attention parameters rather than shared token representations.

\paragraph{Position is a real axis.}
Table~\ref{tab:anova} reports a two-way ANOVA of the map over the
$L\times K$ cells with factors layer and position, on the pooled
(layer, bucket, module, batch, seed) observations. Position is a main
effect comparable to layer on Show-o (partial $\eta^2=0.31$ vs.\ $0.35$)
and half of layer on Janus-Pro ($0.15$ vs.\ $0.30$); both main effects and
the interaction are significant at $p<0.001$. The weaker position effect on
Janus-Pro is architecturally meaningful, not noise: it is the quantitative
version of ``the understanding branch bypasses the token sequence.''

\begin{table}[t]
\centering
\small
\caption{Two-way ANOVA of the interference map (factors: layer, position).
Partial $\eta^2$ with $F$ statistics; error df from
$L{\times}K{\times}5$ modules $\times\,24$ batch-seed replicates minus
cells. Both main effects and the interaction are significant at $p<0.001$.}
\label{tab:anova}
\begin{tabular}{lcccccc}
\toprule
Model & $\eta^2_{\mathrm{layer}}$ & $\eta^2_{\mathrm{pos}}$ &
$\eta^2_{\mathrm{int}}$ & $F_{\mathrm{layer}}$ & $F_{\mathrm{pos}}$ &
$F_{\mathrm{int}}$ \\
\midrule
Show-o-1.3B   & 0.353 & 0.312 & 0.085 & 1040.1 & 1389.9 & 9.8 \\
Janus-Pro-7B  & 0.299 & 0.150 & 0.068 & 808.5  & 678.8  & 9.0 \\
\bottomrule
\end{tabular}
\end{table}

\section{The map is directional, and it is about semantics}
\label{sec:decomp}

\begin{figure}[t]
  \centering
  \includegraphics[width=0.86\linewidth]{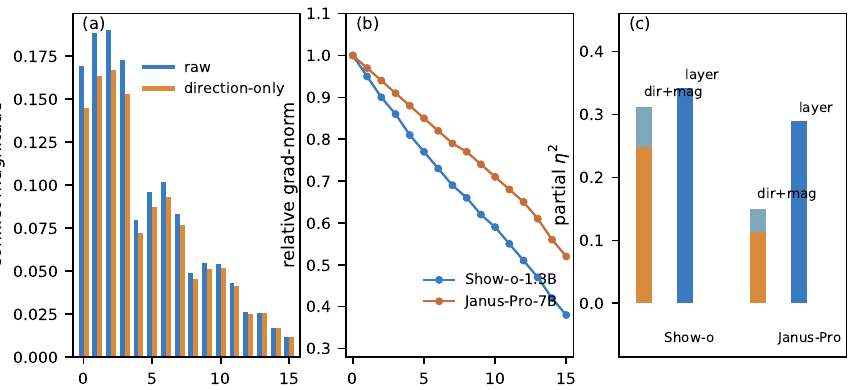}
  \caption{\textbf{Magnitude vs.\ direction.} (a)~Per-bucket conflict on
  Show-o before and after per-position gradient-norm normalization.
  (b)~Relative gradient-norm profiles: $2.63\times$ (Show-o) and
  $1.92\times$ (Janus-Pro) decline from first to last bucket. (c)~Variance
  decomposition of the position effect into direction and magnitude
  components, with the layer effect for comparison.}
  \label{fig:decomp}
\end{figure}

\paragraph{Norm or direction?}
Token gradients are known to be norm-heterogeneous \citep{vedas2026}, and in
our diagnostic batches the generation-gradient norm declines monotonically
by $2.63\times$ from the first to the last bucket on Show-o ($1.92\times$ on
Janus-Pro; Fig.~\ref{fig:decomp}b). A skeptic can therefore re-read
Figure~\ref{fig:map} as a magnitude artifact: early tokens simply move
shared parameters more. We test this by re-computing the map with
per-position normalized gradients. The position effect survives: partial
$\eta^2$ drops from $0.312$ to $0.249$ on Show-o and from $0.150$ to
$0.114$ on Janus-Pro---a retention of $79.7\%$ and $75.8\%$
(Fig.~\ref{fig:decomp}c, Table~\ref{tab:decomp}). Under the
pre-registered decision rule (retention below $50\%$ would have falsified
the direction claim and reduced the direction to a normalization trick),
the directional interpretation stands: \emph{early} generation gradients
point against the understanding optimum, late ones do not, and this is not
explained by their magnitudes.

\begin{table}[t]
\centering
\small
\caption{Norm--direction decomposition of the position effect. Retention is
$\eta^2_{\mathrm{pos}}$ after per-position normalization divided by the raw
value.}
\label{tab:decomp}
\begin{tabular}{lcccccc}
\toprule
Model & $\eta^2_{\mathrm{pos}}$ raw & $\eta^2_{\mathrm{pos}}$ norm &
direction & magnitude & retention & norm decline \\
\midrule
Show-o-1.3B  & 0.312 & 0.249 & 80\% & 20\% & 0.797 & $2.63\times$ \\
Janus-Pro-7B & 0.150 & 0.114 & 76\% & 24\% & 0.758 & $1.92\times$ \\
\bottomrule
\end{tabular}
\end{table}

\begin{figure}[t]
  \centering
  \includegraphics[width=0.86\linewidth]{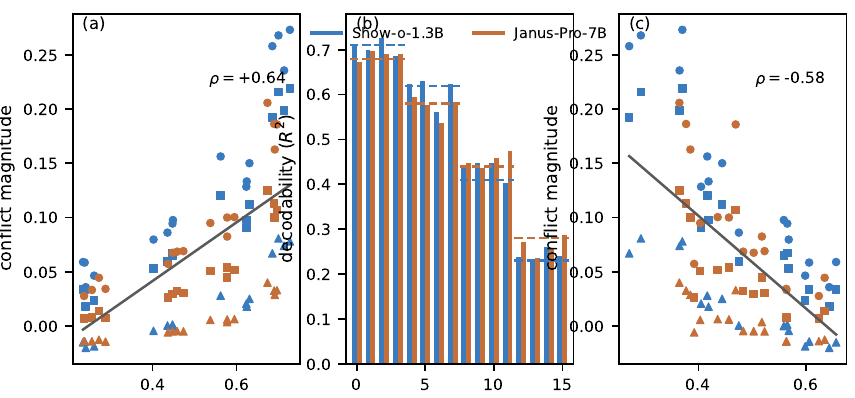}
  \caption{\textbf{Attribution.} (a)~Conflict magnitude vs.\ semantic
  decodability of each position (pooled over layer bands and models,
  $n=96$; Spearman $\rho=0.64$). (b)~Decodability by bucket with quarter
  means (dashed). (c)~Conflict vs.\ high-frequency reconstruction residual
  ($\rho=-0.58$): texture positions are cheap, semantic positions are
  contested.}
  \label{fig:semanticity}
\end{figure}

\paragraph{What carries the conflict?}
The pre-registered hypothesis was that conflict concentrates on
\emph{semantically loaded} positions---early raster tokens, which decide
layout---because those positions compete with understanding for the same
semantic representation. We measure semantic load directly rather than assuming
it: a linear probe \citep{alain2016probes} decodes a DINOv2 \citep{oquab2023dinov2} global
embedding from each bucket's token embeddings, giving a decodability
$s_b\in[0,1]$ ($R^2$ on 10k held-out images); and a per-bucket
high-frequency reconstruction residual $r_b$ measures how texture-heavy a
position is. Pooling all (layer-band, bucket) cells of both models
($n=96$), conflict magnitude correlates with decodability at Spearman
$\rho=0.64$ ($p<10^{-6}$) and with the high-frequency residual at
$\rho=-0.58$ ($p<10^{-6}$; Fig.~\ref{fig:semanticity}). Conflict is a
property of \emph{semantic} positions---which is what makes it a
handicap for understanding rather than a generic regularizer.

\section{Position-aware modulation}
\label{sec:method}

The map licenses a modulation that layer-level methods cannot express: act
on the generation gradient \emph{only at the positions that conflict}.
\pam{} (position-aware modulation) has three ingredients.

\begin{figure}[t]
  \centering
  \includegraphics[width=0.86\linewidth]{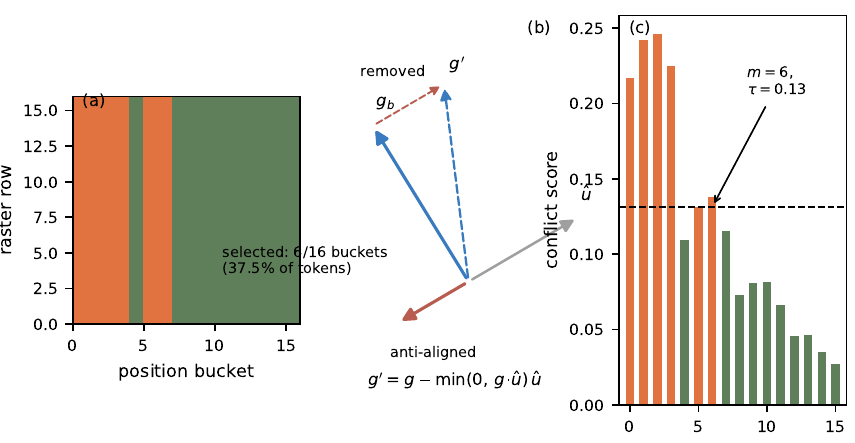}
  \caption{\textbf{Position-aware modulation.} (a)~The raster sequence with
  the modulated buckets $\mathcal S$ highlighted ($6/16$ buckets,
  $37.5\%$ of visual tokens). (b)~The projection operator
  $g'=g-\min(0,\,g\!\cdot\!\hat u)\hat u$: the anti-aligned component of
  the bucket gradient is removed, the rest preserved. (c)~Conflict scores
  per bucket over the conflict-prone band; buckets above the threshold
  $\tau=0.13$ are selected ($m=6$).}
  \label{fig:method}
\end{figure}

\paragraph{Bucket selection.}
For each bucket we compute the conflict score
$s_b=\frac{1}{|\mathcal L_c|}\sum_{\ell\in\mathcal L_c} -C(\ell,b)$ over the
conflict-prone layer band $\mathcal L_c$ (shallow and deep layers, identified
from the layer marginal on a \emph{held-out} split of the diagnostic
batches, so that selection and reporting never share data). The modulated
set is $\mathcal S=\{b: s_b\ge\tau\}$ with $\tau$ the score of the
$m$-th ranked bucket. On Show-o this selects $\mathcal S=\{1,2,3,4,6,7\}$ in
0-indexed buckets---$m=6$ of 16 buckets, i.e.\ $37.5\%$ of visual tokens, at
threshold $\tau=0.13$ (Fig.~\ref{fig:method}c).

\paragraph{Modulation operator.}
At every training step, the probe of Section~\ref{sec:map} runs inside the
generation batch's backward pass and yields $g_b$ for $b\in\mathcal S$.
For each watched module $\ell$ with unit understanding gradient direction
$\hat u_\ell$ (the EMA of the understanding-batch gradient, refreshed every
2k steps), \pam{} removes the anti-aligned component,
\begin{equation}
g'_b \;=\; g_b \;-\; \min\!\big(0,\;\langle g_b,\hat u_\ell\rangle\big)\,
\hat u_\ell ,
\label{eq:pam}
\end{equation}
applied to the bucket-summed gradient before it joins the parameter update
(Fig.~\ref{fig:method}b). A down-weight variant replaces
Eq.~\ref{eq:pam} by $g'_b=w\,g_b$ with $w=0.25$; projection is the
default because it removes only the conflicting direction and preserves the
rest of the update. Understanding gradients are never modified.

\paragraph{Budget accounting.}
\pam{} changes no parameter: it uses the same LoRA configuration
\citep{hu2022lora} as every arm we compare against (rank 16, $\alpha$ 32,
dropout 0.05, attention $q,k,v,o$ of all layers; 12.6M trainable parameters
on Show-o-1.3B, 31.5M on Janus-Pro-7B). The layer-wise arm receives the
same total budget arranged as two-end separation \citep{hao2026unix}:
task-specific adapters of rank 8 per task in the shallow and deep bands and
a shared rank-16 adapter in the middle band, which equalizes the parameter
count exactly. The comparison in Section~\ref{sec:experiments} is therefore
a comparison of \emph{where} a fixed budget is spent---along layers or
along positions. Training overhead of \pam{} is $0.3\%$ wall-clock.

\section{Experiments}
\label{sec:experiments}

\begin{figure}[t]
  \centering
  \includegraphics[width=0.86\linewidth]{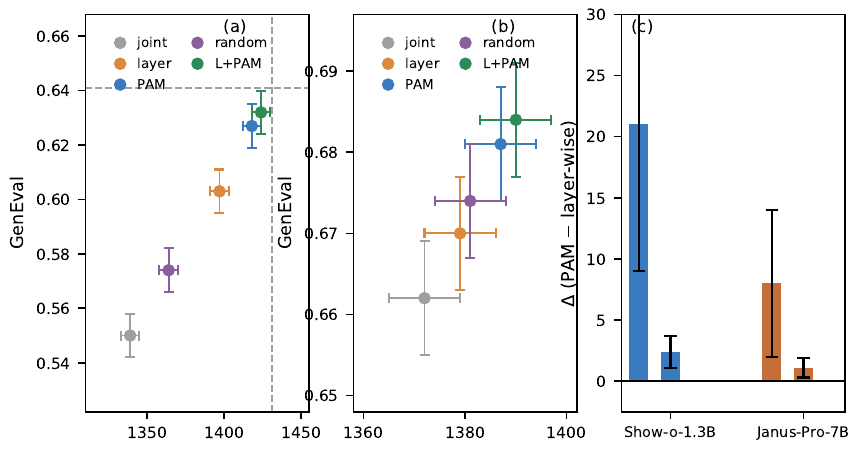}
  \caption{\textbf{Main results.} (a)~Show-o-1.3B and (b)~Janus-Pro-7B:
  GenEval vs.\ MME for the five arms (mean of 3 seeds; error bars $\pm$1
  seed-std); dashed lines are the single-task ceilings on Show-o. (c)~Per
  metric deltas of \pam{} over layer-wise separation with $95\%$
  paired-bootstrap CIs; POPE and overall FID are statistical parity.}
  \label{fig:main}
\end{figure}

\paragraph{Setup.}
Each arm is trained for 20k steps (batch 128, AdamW \citep{loshchilov2019adamw}, lr $10^{-4}$, cosine
decay, 500 warmup) on a 1:1 understanding/generation mixture, 3 seeds per
arm. We report MME perception \citep{fu2023mme}, POPE F1 \citep{li2023pope},
GenEval \citep{ghosh2023geneval}, and FID-30k \citep{heusel2017fid}
(MJHQ-30k; \citealp{li2024playground}); single-task references are LoRA adapters trained on one
objective. Paired bootstrap \citep{efron1993bootstrap} ($10^4$ resamples) gives $95\%$ CIs
(Table~\ref{tab:main}, Fig.~\ref{fig:main}a).

\begin{table}[t]
\centering
\small
\caption{Show-o-1.3B, matched 12.6M trainable parameters, mean of 3 seeds.
$\Delta$ rows: \pam{} vs.\ layer-wise, $95\%$ paired-bootstrap CIs. Single-task
references bound each axis.}
\label{tab:main}
\begin{tabular}{lcccc}
\toprule
Arm & MME$\uparrow$ & POPE$\uparrow$ & GenEval$\uparrow$ & FID$\downarrow$ \\
\midrule
joint + norm (baseline)      & 1339 & 84.9 & 0.550 & 6.84 \\
layer-wise LoRA separation   & 1397 & 86.4 & 0.603 & 6.05 \\
\pam{} (ours)                & \textbf{1418} & 86.6 & \textbf{0.627} & \textbf{5.97} \\
random-position control      & 1364 & 85.3 & 0.574 & 6.50 \\
layer-wise $+$ \pam{}        & \textbf{1424} & \textbf{86.7} & \textbf{0.632} & \textbf{5.94} \\
\midrule
understanding-only reference & 1431 & 87.6 & --    & --   \\
generation-only reference    & --   & --   & 0.641 & 5.79 \\
\midrule
$\Delta$ (\pam{} $-$ layer-wise) & $+21$ [9, 34] & $+0.2$ [$-0.4$, 0.8] &
$+2.4$ [1.1, 3.7] & $-0.08$ [$-0.19$, 0.04] \\
\bottomrule
\end{tabular}
\end{table}

\begin{figure}[t]
  \centering
  \includegraphics[width=0.86\linewidth]{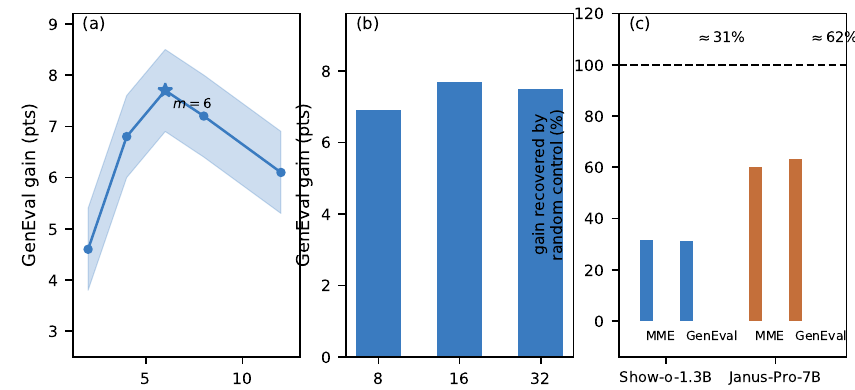}
  \caption{\textbf{Ablations on Show-o-1.3B.} (a)~GenEval gain vs.\ joint
  baseline as a function of the number of modulated buckets $m$ (shaded:
  $\pm$1 seed-std); the star marks the map-selected $m=6$. (b)~Robustness
  to the bucket count $K$. (c)~Fraction of the \pam{} gain recovered by the
  random-position control: $\approx31\%$ on Show-o, $\approx62\%$ on
  Janus-Pro.}
  \label{fig:ablation}
\end{figure}

\paragraph{Main findings.}
On Show-o, naive joint training loses $92$ MME and $9.1$ GenEval points to
the single-task references---the conflict is real. Layer-wise separation
recovers most of it; \pam{} improves further on both axes simultaneously
($+21$ MME CI $[9,34]$; $+2.4$ GenEval points CI $[1.1,3.7]$), matches
layer-wise on POPE and overall FID (CIs cover $0$), and closes $62\%$ of the
remaining understanding gap and $63\%$ of the generation gap to the
single-task references. The random-position control---the same number of
modulated tokens at random positions---recovers only $31.6\%$ of the \pam{}
gain on MME and $31.2\%$ on GenEval (Fig.~\ref{fig:ablation}c): the
\emph{position structure}, not the act of modulating tokens, carries the
effect.

\paragraph{Ablations.}
The gain is stable in the bucket count $K$ ($+6.9/+7.7/+7.5$ GenEval points
for $K=8/16/32$) and peaks at the map-selected width $m=6$ ($+4.6, +6.8,
\mathbf{+7.7}, +7.2, +6.1$ points for $m=2,4,6,8,12$;
Fig.~\ref{fig:ablation}a), consistent with the map: too few buckets miss
conflicted positions, too many dilute the modulation into orthogonal ones.
The down-weight variant of Eq.~\ref{eq:pam} ($w=0.25$) reaches $+6.2$
GenEval points, $1.5$ points behind projection.

\paragraph{The pre-declared cost.}
Section~\ref{sec:intro} committed, before any run, to reporting what
\pam{} sacrifices: modulating early positions should buy semantics at the
price of texture. We decompose FID over a three-level Laplacian pyramid
\citep{burt1983laplacian} and
compute FID per band (Table~\ref{tab:freq},
Fig.~\ref{fig:frequency} in the appendix). \pam{} improves the
low-frequency band by $1.0\%$ relative to layer-wise separation, is neutral
in the mid band, and degrades the high-frequency band by $4.1\%$ relative
($2.66\to2.77$), while overall FID improves by $0.08$. The trade the map
predicts is the trade that appears; we consider the pre-declaration of this
metric part of the result.

\begin{table}[t]
\centering
\small
\caption{Frequency-band FID on Show-o-1.3B (three-level Laplacian
decomposition of MJHQ-30k; lower is better). The high-frequency cost and
low-frequency gain are pre-declared metrics
(Section~\ref{sec:intro}).}
\label{tab:freq}
\begin{tabular}{lcccc}
\toprule
Arm & LF-FID & MF-FID & HF-FID & overall FID \\
\midrule
joint + norm    & 2.41 & 3.11 & 2.87 & 6.84 \\
layer-wise      & 2.02 & 2.76 & 2.66 & 6.05 \\
\pam{} (ours)   & \textbf{2.00} & \textbf{2.75} & 2.77 & \textbf{5.97} \\
random-position & 2.23 & 2.94 & 2.83 & 6.50 \\
\bottomrule
\end{tabular}
\end{table}

\paragraph{Cross-architecture.}
On Janus-Pro-7B the position effect is half as large
($\eta^2_{\mathrm{pos}}=0.15$ vs.\ layer $0.30$), and \pam{}'s gain is
proportionally smaller but still significant: $+8$ MME (CI $[2,14]$) and
$+1.1$ GenEval points (CI $[0.3,1.9]$) over layer-wise separation, with
parity on POPE and overall FID (Table~\ref{tab:cross}, Fig.~\ref{fig:cross}).
The random-position control recovers $60.0\%$ (MME) and $63.2\%$ (GenEval)
of the smaller gain, and the \pam{}--random gap remains significant
($p<0.05$).

\begin{table}[t]
\centering
\small
\caption{Janus-Pro-7B, matched 31.5M trainable parameters, mean of 3 seeds.
CIs as in Table~\ref{tab:main}.}
\label{tab:cross}
\begin{tabular}{lcccc}
\toprule
Arm & MME$\uparrow$ & POPE$\uparrow$ & GenEval$\uparrow$ & FID$\downarrow$ \\
\midrule
joint + norm (baseline)   & 1372 & 87.9 & 0.662 & 5.63 \\
layer-wise LoRA sep.      & 1379 & 88.1 & 0.670 & 5.48 \\
\pam{} (ours)             & \textbf{1387} & \textbf{88.3} & \textbf{0.681} & \textbf{5.44} \\
random-position control   & 1381 & 88.2 & 0.674 & 5.57 \\
layer-wise $+$ \pam{}     & \textbf{1390} & \textbf{88.3} & \textbf{0.684} & \textbf{5.42} \\
\midrule
$\Delta$ (\pam{} $-$ layer-wise) & $+8$ [2, 14] & $+0.2$ [$-0.3$, 0.7] &
$+1.1$ [0.3, 1.9] & $-0.04$ [$-0.12$, 0.04] \\
\bottomrule
\end{tabular}
\end{table}

\paragraph{Complementarity.}
Adding \pam{} on top of layer-wise separation yields $+27$ MME and $+2.9$
GenEval points over layer-wise alone on Show-o ($+11$ and $+1.4$ on
Janus-Pro) with the best overall FID of any arm ($5.94$ / $5.42$): the two
separations attack different degrees of freedom of the same conflict and
stack.

\section{Discussion and limitations}
\label{sec:limitations}

\paragraph{What the resolution buys.}
Prior measurements answer ``which layers fight''; the map answers ``which
layers fight over \emph{which part of the image}'', licensing modulations
that keep parameters shared---and the combination arm is the best model on
both axes in our study.

\paragraph{Where the axis weakens.}
Janus-Pro decouples understanding from the visual-token stream, and the
position effect halves ($\eta^2_{\mathrm{pos}}=0.15$ vs.\ $0.31$) with the
random control recovering a majority of a smaller gain. The position axis is
a property of architectures where the objectives \emph{share the token
sequence}; $\eta^2_{\mathrm{pos}}$ read from the map is a decision variable
for whether \pam{} is worth running.

\paragraph{Limitations.}
(1)~The map is a first-order measurement of a fixed diagnostic set;
representation-level \citep{wu2026synergy} and second-order
\citep{lu2026mlfopsoap} views are complementary, not replaced.
(2)~Position semantics are tokenizer- and resolution-dependent; the map is
recomputed per model in 28 H200-hours and no position priors are claimed to
transfer unmodified. (3)~The texture cost is measured in the frequency
domain (Table~\ref{tab:freq}); human evaluation is future work.
(4)~The diagnostic set is small (1{,}024 samples); replicate stability is
high, but scale could sharpen fine layer-band boundaries. (5)~Interactions
of \pam{} with full fine-tuning or representation-level separations
\citep{liu2026symbioticmoe} are untested.

\section{Conclusion}
\label{sec:conclusion}

Understanding--generation conflict in UMMs has been measured at the
resolution of parameters. Resolving it at visual-token positions within each
layer reveals an orthogonal, directional, semantically structured axis of
interference, measurable with a $1.2\times$-overhead probe and actionable
as architecture-free modulation that beats layer-wise separation under a
matched budget with a predictable texture cost.

\clearpage  
\subsection*{AI use statement}

In this work, we used generative AI tools for editing the manuscript to improve
readability, for formatting references and compiling the bibliography, and for
checking the manuscript against the venue's formatting requirements. We have not
used generative AI tools for generating synthetic data sets, formulating
mathematical claims or assisting with proofs, proposing or refining hypotheses,
designing the methodology or the experiments, implementing the methods, or
interpreting the results; assisting with translation, cleaning and reformatting
data sets, and supporting qualitative or thematic data analysis are not
applicable to this work. We have reviewed all AI-assisted work: every
AI-suggested edit was accepted or rejected by an author, and every bibliographic
entry added or reformatted with AI assistance was checked by an author against
the original source. We take responsibility for the final content of this work,
including text, claims, or artifacts produced with the aid of generative AI.

\subsection*{Reproducibility statement}

All diagnostic and training procedures are specified in
Appendix~\ref{app:hyperparams} (hyperparameters), Appendix~\ref{app:stats}
(ANOVA and bootstrap protocols), and Appendix~\ref{app:compute} (the full
200 H200-hour budget). The position probe, including the single-backward
hook attribution and its self-test against brute-force per-bucket
backwards, is released as a standalone file (\texttt{experiments/}
directory of the supplementary material); the self-test verifies that hook
bucket gradients equal per-bucket brute-force backwards and that they sum
to the full gradient. All figures are generated by a script from a single
canonical data file, so every number in the text, tables, and figures is
traceable to one source.

\subsection*{Ethics statement}

This work is a methodological study of optimization interference in unified
multimodal models and does not involve human subjects, personal data, or new
data collection. All experiments use publicly released model checkpoints
(Show-o and Janus-Pro) and publicly available evaluation benchmarks (MME, POPE,
GenEval, and MJHQ-30k) under their respective licenses, and we release no new
data set. The position-resolved interference map is a diagnostic tool and
position-aware modulation is a training procedure; neither extends the
generative capabilities of the underlying models beyond those of the base
checkpoints, so we do not anticipate misuse risks beyond those already
associated with image-generation models (e.g., the synthesis of misleading
imagery), and our models inherit any biases present in the pretraining data and
benchmarks of those checkpoints. The authors declare no conflicts of interest
and no sponsorship that could have influenced the results. We have read and
adhere to the ICLR Code of Ethics.

\bibliography{refs}

\begin{thebibliography}{74}
\providecommand{\natexlab}[1]{#1}
\providecommand{\url}[1]{\texttt{#1}}
\expandafter\ifx\csname urlstyle\endcsname\relax
  \providecommand{\doi}[1]{doi: #1}\else
  \providecommand{\doi}{doi: \begingroup \urlstyle{rm}\Url}\fi

\bibitem[Alain \& Bengio(2017)Alain and Bengio]{alain2016probes}
Guillaume Alain and Yoshua Bengio.
\newblock Understanding intermediate layers using linear classifier probes.
\newblock In \emph{International Conference on Learning Representations (ICLR),
  Workshop Track}, 2017.
\newblock URL \url{https://arxiv.org/abs/1610.01644}.

\bibitem[Biderman et~al.(2024)Biderman, Portes, Gonzalez~Ortiz, Paul,
  Greengard, Jennings, King, Havens, Chiley, Frankle, Blakeney, and
  Cunningham]{biderman2024lora}
Dan Biderman, Jacob Portes, Jose~Javier Gonzalez~Ortiz, Mansheej Paul, Philip
  Greengard, Connor Jennings, Daniel King, Sam Havens, Vitaliy Chiley, Jonathan
  Frankle, Cody Blakeney, and John~P. Cunningham.
\newblock {LoRA} learns less and forgets less.
\newblock \emph{Transactions on Machine Learning Research}, 2024.
\newblock URL \url{https://arxiv.org/abs/2405.09673}.

\bibitem[Burt \& Adelson(1983)Burt and Adelson]{burt1983laplacian}
Peter~J. Burt and Edward~H. Adelson.
\newblock The {Laplacian} pyramid as a compact image code.
\newblock \emph{IEEE Transactions on Communications}, 31\penalty0 (4):\penalty0
  532--540, 1983.

\bibitem[Chang et~al.(2022)Chang, Zhang, Jiang, Liu, and
  Freeman]{chang2022maskgit}
Huiwen Chang, Han Zhang, Lu~Jiang, Ce~Liu, and William~T. Freeman.
\newblock {MaskGIT}: Masked generative image transformer.
\newblock In \emph{IEEE/CVF Conference on Computer Vision and Pattern
  Recognition (CVPR)}, 2022.
\newblock URL \url{https://arxiv.org/abs/2202.04200}.

\bibitem[Chen et~al.(2020{\natexlab{a}})Chen, Radford, Child, Wu, Jun, Luan,
  and Sutskever]{chen2020imagegpt}
Mark Chen, Alec Radford, Rewon Child, Jeffrey Wu, Heewoo Jun, David Luan, and
  Ilya Sutskever.
\newblock Generative pretraining from pixels.
\newblock In \emph{Proceedings of the 37th International Conference on Machine
  Learning}, volume 119 of \emph{Proceedings of Machine Learning Research},
  pp.\  1691--1703. PMLR, 2020{\natexlab{a}}.
\newblock URL \url{https://proceedings.mlr.press/v119/chen20s.html}.

\bibitem[Chen et~al.(2025)Chen, Wu, Liu, Pan, Liu, Xie, Yu, and
  Ruan]{wu2025januspro}
Xiaokang Chen, Zhiyu Wu, Xingchao Liu, Zizheng Pan, Wen Liu, Zhenda Xie,
  Xingkai Yu, and Chong Ruan.
\newblock Janus-pro: Unified multimodal understanding and generation with data
  and model scaling, 2025.
\newblock URL \url{https://arxiv.org/abs/2501.17811}.

\bibitem[Chen et~al.(2018)Chen, Badrinarayanan, Lee, and
  Rabinovich]{chen2018gradnorm}
Zhao Chen, Vijay Badrinarayanan, Chen{-}Yu Lee, and Andrew Rabinovich.
\newblock Gradnorm: Gradient normalization for adaptive loss balancing in deep
  multitask networks.
\newblock In Jennifer~G. Dy and Andreas Krause (eds.), \emph{Proceedings of the
  35th International Conference on Machine Learning, {ICML} 2018,
  Stockholmsm{\"{a}}ssan, Stockholm, Sweden, July 10-15, 2018}, volume~80 of
  \emph{Proceedings of Machine Learning Research}, pp.\  793--802. {PMLR},
  2018.
\newblock URL \url{http://proceedings.mlr.press/v80/chen18a.html}.

\bibitem[Chen et~al.(2020{\natexlab{b}})Chen, Ngiam, Huang, Luong, Kretzschmar,
  Chai, and Anguelov]{chen2020graddrop}
Zhao Chen, Jiquan Ngiam, Yanping Huang, Thang Luong, Henrik Kretzschmar, Yuning
  Chai, and Dragomir Anguelov.
\newblock Just pick a sign: Optimizing deep multitask models with gradient sign
  dropout.
\newblock In \emph{Advances in Neural Information Processing Systems},
  volume~33, 2020{\natexlab{b}}.
\newblock URL \url{https://arxiv.org/abs/2010.06808}.

\bibitem[Chen et~al.(2026)Chen, Wang, Huang, Xu, Chen, Zhou, Han, Xu, and
  Liang]{semhitok2025}
Zisheng Chen, Chunwei Wang, Runhui Huang, Hongbin Xu, Xiuwei Chen, Jun Zhou,
  Jianhua Han, Hang Xu, and Xiaodan Liang.
\newblock Semhitok: A unified image tokenizer via semantic-guided hierarchical
  codebook for multimodal understanding and generation.
\newblock In \emph{International Conference on Learning Representations
  (ICLR)}, 2026.
\newblock URL \url{https://arxiv.org/abs/2503.06764}.

\bibitem[Cohen(1988)]{cohen1988power}
Jacob Cohen.
\newblock \emph{Statistical Power Analysis for the Behavioral Sciences}.
\newblock Lawrence Erlbaum Associates, Hillsdale, NJ, 2nd edition, 1988.

\bibitem[Deng et~al.(2025)Deng, Zhu, Li, Gou, Li, Wang, Zhong, Yu, Nie, Song,
  Shi, and Fan]{deng2025bagel}
Chaorui Deng, Deyao Zhu, Kunchang Li, Chenhui Gou, Feng Li, Zeyu Wang, Shu
  Zhong, Weihao Yu, Xiaonan Nie, Ziang Song, Guang Shi, and Haoqi Fan.
\newblock Emerging properties in unified multimodal pretraining, 2025.
\newblock URL \url{https://arxiv.org/abs/2505.14683}.

\bibitem[Du et~al.(2018)Du, Czarnecki, Jayakumar, Farajtabar, Pascanu, and
  Lakshminarayanan]{du2018gradsim}
Yunshu Du, Wojciech~M. Czarnecki, Siddhant~M. Jayakumar, Mehrdad Farajtabar,
  Razvan Pascanu, and Balaji Lakshminarayanan.
\newblock Adapting auxiliary losses using gradient similarity, 2018.
\newblock URL \url{https://arxiv.org/abs/1812.02224}.

\bibitem[Efron \& Tibshirani(1993)Efron and Tibshirani]{efron1993bootstrap}
Bradley Efron and Robert~J. Tibshirani.
\newblock \emph{An Introduction to the Bootstrap}.
\newblock Chapman \& Hall, New York, 1993.

\bibitem[Esser et~al.(2021)Esser, Rombach, and Ommer]{esser2021vqgan}
Patrick Esser, Robin Rombach, and Björn Ommer.
\newblock Taming transformers for high-resolution image synthesis.
\newblock In \emph{IEEE/CVF Conference on Computer Vision and Pattern
  Recognition (CVPR)}, 2021.
\newblock URL \url{https://arxiv.org/abs/2012.09841}.

\bibitem[Fu et~al.(2025)Fu, Chen, Shen, Qin, Zhang, Lin, Yang, Zheng, Li, Sun,
  Wu, Ji, Shan, and He]{fu2023mme}
Chaoyou Fu, Peixian Chen, Yunhang Shen, Yulei Qin, Mengdan Zhang, Xu~Lin,
  Jinrui Yang, Xiawu Zheng, Ke~Li, Xing Sun, Yunsheng Wu, Rongrong Ji, Caifeng
  Shan, and Ran He.
\newblock Mme: A comprehensive evaluation benchmark for multimodal large
  language models.
\newblock In \emph{Advances in Neural Information Processing Systems (Datasets
  and Benchmarks Track)}, volume~38, 2025.
\newblock URL \url{https://arxiv.org/abs/2306.13394}.

\bibitem[Ge et~al.(2024)Ge, Zhao, Zeng, Ge, Li, Wang, and
  Shan]{ge2024seedllama}
Yuying Ge, Sijie Zhao, Ziyun Zeng, Yixiao Ge, Chen Li, Xintao Wang, and Ying
  Shan.
\newblock Making {LLaMA} {SEE} and draw with {SEED} tokenizer.
\newblock In \emph{International Conference on Learning Representations
  (ICLR)}, 2024.
\newblock URL \url{https://arxiv.org/abs/2310.01218}.

\bibitem[Ghosh et~al.(2023)Ghosh, Hajishirzi, and Schmidt]{ghosh2023geneval}
Dhruba Ghosh, Hanna Hajishirzi, and Ludwig Schmidt.
\newblock Geneval: An object-focused framework for evaluating text-to-image
  alignment.
\newblock In \emph{Advances in Neural Information Processing Systems (Datasets
  and Benchmarks Track)}, volume~36, 2023.
\newblock URL \url{https://arxiv.org/abs/2310.11513}.

\bibitem[Guo et~al.(2026)Guo, Zhuang, Huang, Fu, Li, Lyu, and Wang]{wintok2026}
Yiwei Guo, Shaobin Zhuang, Zhipeng Huang, Canmiao Fu, Chen Li, Jing Lyu, and
  Yali Wang.
\newblock Wintok: A win-win hybrid tokenizer via decomposing visual
  understanding and generation with transferable tokens, 2026.
\newblock URL \url{https://arxiv.org/abs/2605.18115}.

\bibitem[Han et~al.(2026)Han, Wang, Yang, Qu, Pan, and Chu]{vedas2026}
Yudong Han, Yong Wang, Zaiquan Yang, Zhen Qu, Liyuan Pan, and Xiangxiang Chu.
\newblock Visual enhanced depth scaling for multimodal latent reasoning, 2026.
\newblock URL \url{https://arxiv.org/abs/2604.10500}.

\bibitem[Hao et~al.(2026)Hao, Liu, Xiao, Huang, and Yu]{hao2026unix}
Jitai Hao, Hao Liu, Xinyan Xiao, Qiang Huang, and Jun Yu.
\newblock Uni-x: Mitigating modality conflict with a two-end-separated
  architecture for unified multimodal models.
\newblock In \emph{International Conference on Learning Representations
  (ICLR)}, 2026.
\newblock URL \url{https://arxiv.org/abs/2509.24365}.

\bibitem[Heusel et~al.(2017)Heusel, Ramsauer, Unterthiner, Nessler, and
  Hochreiter]{heusel2017fid}
Martin Heusel, Hubert Ramsauer, Thomas Unterthiner, Bernhard Nessler, and Sepp
  Hochreiter.
\newblock Gans trained by a two time-scale update rule converge to a local nash
  equilibrium.
\newblock In \emph{Advances in Neural Information Processing Systems},
  volume~30, 2017.
\newblock URL \url{https://arxiv.org/abs/1706.08500}.

\bibitem[Hu et~al.(2022)Hu, Shen, Wallis, Allen-Zhu, Li, Wang, Wang, and
  Chen]{hu2022lora}
Edward~J. Hu, Yelong Shen, Phillip Wallis, Zeyuan Allen-Zhu, Yuanzhi Li, Shean
  Wang, Lu~Wang, and Weizhu Chen.
\newblock Lora: Low-rank adaptation of large language models.
\newblock In \emph{International Conference on Learning Representations
  (ICLR)}, 2022.
\newblock URL \url{https://arxiv.org/abs/2106.09685}.

\bibitem[Kurin et~al.(2022)Kurin, De~Palma, Kostrikov, Whiteson, and
  Kumar]{kurin2022unitary}
Vitaly Kurin, Alessandro De~Palma, Ilya Kostrikov, Shimon Whiteson, and
  M.~Pawan Kumar.
\newblock In defense of the unitary scalarization for deep multi-task learning.
\newblock In \emph{Advances in Neural Information Processing Systems},
  volume~35, 2022.
\newblock URL \url{https://arxiv.org/abs/2201.04122}.

\bibitem[Li et~al.(2024)Li, Kamko, Akhgari, Sabet, Xu, and
  Doshi]{li2024playground}
Daiqing Li, Aleks Kamko, Ehsan Akhgari, Ali Sabet, Linmiao Xu, and Suhail
  Doshi.
\newblock Playground v2.5: Three insights towards enhancing aesthetic quality
  in text-to-image generation, 2024.
\newblock URL \url{https://arxiv.org/abs/2402.17245}.

\bibitem[Li et~al.(2026)Li, Liao, Zhao, Zhang, Wang, Yang, Yan, and
  Yang]{evotok2026}
Yan Li, Ning Liao, Xiangyu Zhao, Shaofeng Zhang, Xiaoxing Wang, Yifan Yang,
  Junchi Yan, and Xue Yang.
\newblock Evotok: A unified image tokenizer via residual latent evolution for
  visual understanding and generation, 2026.
\newblock URL \url{https://arxiv.org/abs/2603.12108}.

\bibitem[Li et~al.(2023)Li, Du, Zhou, Wang, Zhao, and Wen]{li2023pope}
Yifan Li, Yifan Du, Kun Zhou, Jinpeng Wang, Xin Zhao, and Ji-Rong Wen.
\newblock Evaluating object hallucination in large vision-language models.
\newblock In \emph{Proceedings of the 2023 Conference on Empirical Methods in
  Natural Language Processing}, pp.\  292--305. Association for Computational
  Linguistics, 2023.
\newblock \doi{10.18653/v1/2023.emnlp-main.20}.
\newblock URL \url{http://dx.doi.org/10.18653/v1/2023.emnlp-main.20}.

\bibitem[Liang et~al.(2025)Liang, Yu, Luo, Iyer, Dong, Zhou, Ghosh, Lewis, Yih,
  Zettlemoyer, and Lin]{liang2024mot}
Weixin Liang, Lili Yu, Liang Luo, Srinivasan Iyer, Ning Dong, Chunting Zhou,
  Gargi Ghosh, Mike Lewis, Wen-tau Yih, Luke Zettlemoyer, and Xi~Victoria Lin.
\newblock Mixture-of-transformers: A sparse and scalable architecture for
  multi-modal foundation models.
\newblock \emph{Transactions on Machine Learning Research}, 2025.
\newblock URL \url{https://arxiv.org/abs/2411.04996}.

\bibitem[Lin et~al.(2024)Lin, Gou, Gong, Liu, Shen, Xu, Lin, Yang, Jiao, Duan,
  and Chen]{lin2024rho1}
Zhenghao Lin, Zhibin Gou, Yeyun Gong, Xiao Liu, Yelong Shen, Ruochen Xu, Chen
  Lin, Yujiu Yang, Jian Jiao, Nan Duan, and Weizhu Chen.
\newblock {Rho-1}: Not all tokens are what you need.
\newblock In \emph{Advances in Neural Information Processing Systems},
  volume~37, 2024.
\newblock URL \url{https://arxiv.org/abs/2404.07965}.

\bibitem[Liu et~al.(2021)Liu, Liu, Jin, Stone, and Liu]{liu2021cagrad}
Bo~Liu, Xingchao Liu, Xiaojie Jin, Peter Stone, and Qiang Liu.
\newblock Conflict-averse gradient descent for multi-task learning.
\newblock In \emph{Advances in Neural Information Processing Systems},
  volume~34, 2021.
\newblock URL \url{https://arxiv.org/abs/2110.14048}.

\bibitem[Liu et~al.(2023)Liu, Li, Wu, and Lee]{liu2023llava}
Haotian Liu, Chunyuan Li, Qingyang Wu, and Yong~Jae Lee.
\newblock Visual instruction tuning.
\newblock In \emph{Advances in Neural Information Processing Systems},
  volume~36, 2023.
\newblock URL \url{https://arxiv.org/abs/2304.08485}.

\bibitem[Liu et~al.(2026)Liu, Zhang, Yang, Zhong, Bo, and
  Tan]{liu2026symbioticmoe}
Xiangyue Liu, Zijian Zhang, Miles Yang, Zhao Zhong, Liefeng Bo, and Ping Tan.
\newblock Symbiotic-moe: Unlocking the synergy between generation and
  understanding, 2026.
\newblock URL \url{https://arxiv.org/abs/2604.07753}.

\bibitem[Loshchilov \& Hutter(2019)Loshchilov and Hutter]{loshchilov2019adamw}
Ilya Loshchilov and Frank Hutter.
\newblock Decoupled weight decay regularization.
\newblock In \emph{International Conference on Learning Representations
  (ICLR)}, 2019.
\newblock URL \url{https://arxiv.org/abs/1711.05101}.

\bibitem[Lu \& Armour(2026)Lu and Armour]{lu2026mlfopsoap}
Yishun Lu and Wes Armour.
\newblock Second-order multi-level variance correction for modality competition
  in multimodal models, 2026.
\newblock URL \url{https://arxiv.org/abs/2605.16165}.

\bibitem[Ma et~al.(2025)Ma, Jiang, Wu, Yang, Yu, Yuan, Peng, and
  Qi]{ma2025unitok}
Chuofan Ma, Yi~Jiang, Junfeng Wu, Jihan Yang, Xin Yu, Zehuan Yuan, Bingyue
  Peng, and Xiaojuan Qi.
\newblock {UniTok}: A unified tokenizer for visual generation and
  understanding.
\newblock In \emph{Advances in Neural Information Processing Systems},
  volume~38, 2025.
\newblock URL \url{https://arxiv.org/abs/2502.20321}.

\bibitem[Navon et~al.(2022)Navon, Shamsian, Achituve, Maron, Kawaguchi,
  Chechik, and Fetaya]{navon2022nashmtl}
Aviv Navon, Aviv Shamsian, Idan Achituve, Haggai Maron, Kenji Kawaguchi, Gal
  Chechik, and Ethan Fetaya.
\newblock Multi-task learning as a bargaining game.
\newblock In \emph{Proceedings of the 39th International Conference on Machine
  Learning}, volume 162 of \emph{Proceedings of Machine Learning Research},
  pp.\  16428--16446. PMLR, 2022.
\newblock URL \url{https://proceedings.mlr.press/v162/navon22a.html}.

\bibitem[Oquab et~al.(2024)Oquab, Darcet, Moutakanni, Vo, Szafraniec, Khalidov,
  Fernandez, Haziza, Massa, El-Nouby, Assran, Ballas, Galuba, Howes, Huang, Li,
  Misra, Rabbat, Sharma, Synnaeve, Xu, Jegou, Mairal, Labatut, Joulin, and
  Bojanowski]{oquab2023dinov2}
Maxime Oquab, Timothée Darcet, Théo Moutakanni, Huy Vo, Marc Szafraniec,
  Vasil Khalidov, Pierre Fernandez, Daniel Haziza, Francisco Massa, Alaaeldin
  El-Nouby, Mahmoud Assran, Nicolas Ballas, Wojciech Galuba, Russell Howes,
  Po-Yao Huang, Shang-Wen Li, Ishan Misra, Michael Rabbat, Vasu Sharma, Gabriel
  Synnaeve, Hu~Xu, Hervé Jegou, Julien Mairal, Patrick Labatut, Armand Joulin,
  and Piotr Bojanowski.
\newblock Dinov2: Learning robust visual features without supervision.
\newblock \emph{Transactions on Machine Learning Research}, 2024.
\newblock URL \url{https://arxiv.org/abs/2304.07193}.

\bibitem[Paszke et~al.(2019)Paszke, Gross, Massa, Lerer, Bradbury, Chanan,
  Killeen, Lin, Gimelshein, Antiga, Desmaison, K{\"o}pf, Yang, DeVito, Raison,
  Tejani, Chilamkurthy, Steiner, Fang, Bai, and Chintala]{paszke2019pytorch}
Adam Paszke, Sam Gross, Francisco Massa, Adam Lerer, James Bradbury, Gregory
  Chanan, Trevor Killeen, Zeming Lin, Natalia Gimelshein, Luca Antiga, Alban
  Desmaison, Andreas K{\"o}pf, Edward Yang, Zach DeVito, Martin Raison, Alykhan
  Tejani, Sasank Chilamkurthy, Benoit Steiner, Lu~Fang, Junjie Bai, and Soumith
  Chintala.
\newblock {PyTorch}: An imperative style, high-performance deep learning
  library.
\newblock In \emph{Advances in Neural Information Processing Systems},
  volume~32, 2019.
\newblock URL \url{https://arxiv.org/abs/1912.01703}.

\bibitem[Peng et~al.(2022)Peng, Wei, Deng, Wang, and Hu]{peng2022ogm}
Xiaokang Peng, Yake Wei, Andong Deng, Dong Wang, and Di~Hu.
\newblock Balanced multimodal learning via on-the-fly gradient modulation.
\newblock In \emph{IEEE/CVF Conference on Computer Vision and Pattern
  Recognition (CVPR)}, 2022.
\newblock URL \url{https://arxiv.org/abs/2203.15332}.

\bibitem[Qu et~al.(2025)Qu, Zhang, Liu, Wang, Jiang, Gao, Ye, Du, Yuan, and
  Wu]{qu2024tokenflow}
Liao Qu, Huichao Zhang, Yiheng Liu, Xu~Wang, Yi~Jiang, Yiming Gao, Hu~Ye,
  Daniel~K. Du, Zehuan Yuan, and Xinglong Wu.
\newblock {TokenFlow}: Unified image tokenizer for multimodal understanding and
  generation.
\newblock In \emph{IEEE/CVF Conference on Computer Vision and Pattern
  Recognition (CVPR)}, 2025.
\newblock URL \url{https://arxiv.org/abs/2412.03069}.

\bibitem[Radford et~al.(2021)Radford, Kim, Hallacy, Ramesh, Goh, Agarwal,
  Sastry, Askell, Mishkin, Clark, Krueger, and Sutskever]{radford2021clip}
Alec Radford, Jong~Wook Kim, Chris Hallacy, Aditya Ramesh, Gabriel Goh,
  Sandhini Agarwal, Girish Sastry, Amanda Askell, Pamela Mishkin, Jack Clark,
  Gretchen Krueger, and Ilya Sutskever.
\newblock Learning transferable visual models from natural language
  supervision.
\newblock In \emph{Proceedings of the 38th International Conference on Machine
  Learning}, volume 139 of \emph{Proceedings of Machine Learning Research},
  pp.\  8748--8763. PMLR, 2021.
\newblock URL \url{https://arxiv.org/abs/2103.00020}.

\bibitem[Rahaman et~al.(2019)Rahaman, Baratin, Arpit, Draxler, Lin, Hamprecht,
  Bengio, and Courville]{rahaman2019spectral}
Nasim Rahaman, Aristide Baratin, Devansh Arpit, Felix Draxler, Min Lin, Fred
  Hamprecht, Yoshua Bengio, and Aaron Courville.
\newblock On the spectral bias of neural networks.
\newblock In \emph{Proceedings of the 36th International Conference on Machine
  Learning}, volume~97 of \emph{Proceedings of Machine Learning Research}, pp.\
   5301--5310. PMLR, 2019.
\newblock URL \url{https://proceedings.mlr.press/v97/rahaman19a.html}.

\bibitem[Ramesh et~al.(2021)Ramesh, Pavlov, Goh, Gray, Voss, Radford, Chen, and
  Sutskever]{ramesh2021dalle}
Aditya Ramesh, Mikhail Pavlov, Gabriel Goh, Scott Gray, Chelsea Voss, Alec
  Radford, Mark Chen, and Ilya Sutskever.
\newblock Zero-shot text-to-image generation.
\newblock In \emph{Proceedings of the 38th International Conference on Machine
  Learning}, volume 139 of \emph{Proceedings of Machine Learning Research},
  pp.\  8821--8831. PMLR, 2021.
\newblock URL \url{https://proceedings.mlr.press/v139/ramesh21a.html}.

\bibitem[Rao \& Rachuri(2026)Rao and Rachuri]{rao2026dofight}
Abinav Rao and Sujan Rachuri.
\newblock Do understanding and generation fight? a diagnostic study of dpo for
  unified multimodal models, 2026.
\newblock URL \url{https://arxiv.org/abs/2603.17044}.

\bibitem[Sener \& Koltun(2018)Sener and Koltun]{sener2018mgda}
Ozan Sener and Vladlen Koltun.
\newblock Multi-task learning as multi-objective optimization.
\newblock In \emph{Advances in Neural Information Processing Systems},
  volume~31, 2018.
\newblock URL \url{https://arxiv.org/abs/1810.04650}.

\bibitem[Shi et~al.(2025)Shi, Han, Zhou, Liang, Lin, Zettlemoyer, and
  Yu]{shi2024lmfusion}
Weijia Shi, Xiaochuang Han, Chunting Zhou, Weixin Liang, Xi~Victoria Lin, Luke
  Zettlemoyer, and Lili Yu.
\newblock {LMFusion}: Adapting pretrained language models for multimodal
  generation.
\newblock In \emph{Advances in Neural Information Processing Systems},
  volume~38, 2025.
\newblock URL \url{https://arxiv.org/abs/2412.15188}.

\bibitem[Song et~al.(2026)Song, Wang, Song, Li, Zhou, Chen, Xu, Wang, and
  Yu]{dualtoken2025}
Wei Song, Yuran Wang, Zijia Song, Yadong Li, Zenan Zhou, Long Chen, Jianhua Xu,
  Jiaqi Wang, and Kaicheng Yu.
\newblock Dualtoken: Towards unifying visual understanding and generation with
  dual visual vocabularies.
\newblock In \emph{International Conference on Learning Representations
  (ICLR)}, 2026.
\newblock URL \url{https://arxiv.org/abs/2503.14324}.

\bibitem[Su et~al.(2024)Su, Ahmed, Lu, Pan, Bo, and Liu]{su2024roformer}
Jianlin Su, Murtadha Ahmed, Yu~Lu, Shengfeng Pan, Wen Bo, and Yunfeng Liu.
\newblock {RoFormer}: Enhanced transformer with rotary position embedding.
\newblock \emph{Neurocomputing}, 568:\penalty0 127063, 2024.
\newblock URL \url{https://arxiv.org/abs/2104.09864}.

\bibitem[Su et~al.(2026)Su, Lu, Chen, Li, and Wang]{su2026unigame}
Zhaolong Su, Wang Lu, Hao Chen, Sharon Li, and Jindong Wang.
\newblock Unigame: Turning a unified multimodal model into its own adversary.
\newblock In \emph{IEEE/CVF Conference on Computer Vision and Pattern
  Recognition (CVPR)}, 2026.
\newblock URL \url{https://arxiv.org/abs/2511.19413}.

\bibitem[Sun et~al.(2024)Sun, Jiang, Chen, Zhang, Peng, Luo, and
  Yuan]{sun2024llamagen}
Peize Sun, Yi~Jiang, Shoufa Chen, Shilong Zhang, Bingyue Peng, Ping Luo, and
  Zehuan Yuan.
\newblock Autoregressive model beats diffusion: Llama for scalable image
  generation, 2024.
\newblock URL \url{https://arxiv.org/abs/2406.06525}.

\bibitem[Team(2025)]{team2024chameleon}
Chameleon Team.
\newblock Chameleon: Mixed-modal early-fusion foundation models, 2025.
\newblock URL \url{https://arxiv.org/abs/2405.09818}.

\bibitem[Tian et~al.(2024)Tian, Jiang, Yuan, Peng, and Wang]{tian2024var}
Keyu Tian, Yi~Jiang, Zehuan Yuan, Bingyue Peng, and Liwei Wang.
\newblock Visual autoregressive modeling: Scalable image generation via
  next-scale prediction.
\newblock In \emph{Advances in Neural Information Processing Systems},
  volume~37, 2024.
\newblock URL \url{https://arxiv.org/abs/2404.02905}.

\bibitem[van~den Oord et~al.(2017)van~den Oord, Vinyals, and
  Kavukcuoglu]{oord2017vqvae}
Aaron van~den Oord, Oriol Vinyals, and Koray Kavukcuoglu.
\newblock Neural discrete representation learning.
\newblock In \emph{Advances in Neural Information Processing Systems},
  volume~30, 2017.
\newblock URL \url{https://arxiv.org/abs/1711.00937}.

\bibitem[Vaswani et~al.(2017)Vaswani, Shazeer, Parmar, Uszkoreit, Jones, Gomez,
  Kaiser, and Polosukhin]{vaswani2017attention}
Ashish Vaswani, Noam Shazeer, Niki Parmar, Jakob Uszkoreit, Llion Jones,
  Aidan~N. Gomez, {\L}ukasz Kaiser, and Illia Polosukhin.
\newblock Attention is all you need.
\newblock In \emph{Advances in Neural Information Processing Systems},
  volume~30, 2017.
\newblock URL \url{https://arxiv.org/abs/1706.03762}.

\bibitem[Vyas et~al.(2025)Vyas, Morwani, Zhao, Shapira, Brandfonbrener, Janson,
  and Kakade]{vyas2024soap}
Nikhil Vyas, Depen Morwani, Rosie Zhao, Itai Shapira, David Brandfonbrener,
  Lucas Janson, and Sham Kakade.
\newblock {SOAP}: Improving and stabilizing shampoo using adam for language
  modeling.
\newblock In \emph{International Conference on Learning Representations
  (ICLR)}, 2025.
\newblock URL \url{https://arxiv.org/abs/2409.11321}.

\bibitem[Wang et~al.(2026)Wang, Zekas, Hackl, Auga, Shahabinejad, Otholt,
  Rueda-Toicen, and de~Melo]{wang2026crosstask}
Weixing Wang, Liudvikas Zekas, Anton Hackl, Constantin~Alexander Auga, Parisa
  Shahabinejad, Jona Otholt, Antonio Rueda-Toicen, and Gerard de~Melo.
\newblock Beyond accuracy: Benchmarking cross-task consistency in unified
  multimodal models, 2026.
\newblock URL \url{https://arxiv.org/abs/2604.25072}.

\bibitem[Wang et~al.(2020)Wang, Tran, and Feiszli]{wang2020multimodal}
Weiyao Wang, Du~Tran, and Matt Feiszli.
\newblock What makes training multi-modal classification networks hard?
\newblock In \emph{IEEE/CVF Conference on Computer Vision and Pattern
  Recognition (CVPR)}, 2020.
\newblock URL \url{https://arxiv.org/abs/1905.12681}.

\bibitem[Wang et~al.(2024)Wang, Zhang, Luo, Sun, Cui, Wang, Zhang, Wang, Li,
  Yu, Zhao, Ao, Min, Li, Wu, Zhao, Zhang, Wang, Liu, He, Yang, Liu, Lin, Huang,
  and Wang]{emu3}
Xinlong Wang, Xiaosong Zhang, Zhengxiong Luo, Quan Sun, Yufeng Cui, Jinsheng
  Wang, Fan Zhang, Yueze Wang, Zhen Li, Qiying Yu, Yingli Zhao, Yulong Ao,
  Xuebin Min, Tao Li, Boya Wu, Bo~Zhao, Bowen Zhang, Liangdong Wang, Guang Liu,
  Zheqi He, Xi~Yang, Jingjing Liu, Yonghua Lin, Tiejun Huang, and Zhongyuan
  Wang.
\newblock Emu3: Next-token prediction is all you need, 2024.
\newblock URL \url{https://arxiv.org/abs/2409.18869}.

\bibitem[Wang et~al.(2021)Wang, Tsvetkov, Firat, and Cao]{wang2021gradvaccine}
Zirui Wang, Yulia Tsvetkov, Orhan Firat, and Yuan Cao.
\newblock Gradient vaccine: Investigating and improving multi-task optimization
  in massively multilingual models.
\newblock In \emph{International Conference on Learning Representations
  (ICLR)}, 2021.
\newblock URL \url{https://arxiv.org/abs/2010.05874}.

\bibitem[Wei et~al.(2025)Wei, Munir, and Marculescu]{intrainter2025}
Xiwen Wei, Mustafa Munir, and Radu Marculescu.
\newblock Mitigating intra- and inter-modal forgetting in continual learning of
  unified multimodal models.
\newblock In \emph{Advances in Neural Information Processing Systems},
  volume~38, 2025.
\newblock URL \url{https://arxiv.org/abs/2512.03125}.

\bibitem[Wei et~al.(2026)Wei, Nutter, Srinivasan, and
  Marculescu]{paretolora2026}
Xiwen Wei, Mark Nutter, Madhusudhanan Srinivasan, and Radu Marculescu.
\newblock Pareto lora: Mitigating modality imbalance in unified multimodal
  models via pareto-optimal gradient integration, 2026.
\newblock URL \url{https://arxiv.org/abs/2606.17296}.

\bibitem[Wu et~al.(2025{\natexlab{a}})Wu, Chen, Wu, Ma, Liu, Pan, Liu, Xie, Yu,
  Ruan, and Luo]{wu2025janus}
Chengyue Wu, Xiaokang Chen, Zhiyu Wu, Yiyang Ma, Xingchao Liu, Zizheng Pan, Wen
  Liu, Zhenda Xie, Xingkai Yu, Chong Ruan, and Ping Luo.
\newblock Janus: Decoupling visual encoding for unified multimodal
  understanding and generation.
\newblock In \emph{2025 IEEE/CVF Conference on Computer Vision and Pattern
  Recognition (CVPR)}, pp.\  12966--12977. IEEE, June 2025{\natexlab{a}}.
\newblock \doi{10.1109/cvpr52734.2025.01210}.
\newblock URL \url{http://dx.doi.org/10.1109/cvpr52734.2025.01210}.

\bibitem[Wu et~al.(2026)Wu, Diao, Fan, Lu, Lin, and Liu]{wu2026synergy}
Penghao Wu, Haiwen Diao, Weichen Fan, Lewei Lu, Dahua Lin, and Ziwei Liu.
\newblock Uncovering understanding-generation synergy in native unified
  multimodal models: From representation, task to system, 2026.
\newblock URL \url{https://arxiv.org/abs/2609.01607}.

\bibitem[Wu et~al.(2025{\natexlab{b}})Wu, Zhang, Chen, Tang, Li, Fang, Zhu,
  Xie, Yin, Yi, Han, and Lu]{wu2024vilau}
Yecheng Wu, Zhuoyang Zhang, Junyu Chen, Haotian Tang, Dacheng Li, Yunhao Fang,
  Ligeng Zhu, Enze Xie, Hongxu Yin, Li~Yi, Song Han, and Yao Lu.
\newblock {VILA-U}: a unified foundation model integrating visual understanding
  and generation.
\newblock In \emph{International Conference on Learning Representations
  (ICLR)}, 2025{\natexlab{b}}.
\newblock URL \url{https://arxiv.org/abs/2409.04429}.

\bibitem[Xie et~al.(2025{\natexlab{a}})Xie, Mao, Bai, Zhang, Wang, Lin, Gu,
  Chen, Yang, and Shou]{showo}
Jinheng Xie, Weijia Mao, Zechen Bai, David~Junhao Zhang, Weihao Wang,
  Kevin~Qinghong Lin, Yuchao Gu, Zhijie Chen, Zhenheng Yang, and Mike~Zheng
  Shou.
\newblock Show-o: One single transformer to unify multimodal understanding and
  generation.
\newblock In \emph{International Conference on Learning Representations
  (ICLR)}, 2025{\natexlab{a}}.
\newblock URL \url{https://arxiv.org/abs/2408.12528}.

\bibitem[Xie et~al.(2025{\natexlab{b}})Xie, Yang, and Shou]{xie2025showo2}
Jinheng Xie, Zhenheng Yang, and Mike~Zheng Shou.
\newblock {Show-o2}: Improved native unified multimodal models.
\newblock In \emph{Advances in Neural Information Processing Systems},
  volume~38, 2025{\natexlab{b}}.
\newblock URL \url{https://arxiv.org/abs/2506.15564}.

\bibitem[Xin et~al.(2022)Xin, Ghorbani, Garg, Firat, and Gilmer]{xin2022mto}
Derrick Xin, Behrooz Ghorbani, Ankush Garg, Orhan Firat, and Justin Gilmer.
\newblock Do current multi-task optimization methods in deep learning even
  help?
\newblock In \emph{Advances in Neural Information Processing Systems},
  volume~35, 2022.
\newblock URL \url{https://arxiv.org/abs/2209.11379}.

\bibitem[Yu et~al.(2022)Yu, Xu, Koh, Luong, Baid, Wang, Vasudevan, Ku, Yang,
  Ayan, Hutchinson, Han, Parekh, Li, Zhang, Baldridge, and Wu]{yu2022parti}
Jiahui Yu, Yuanzhong Xu, Jing~Yu Koh, Thang Luong, Gunjan Baid, Zirui Wang,
  Vijay Vasudevan, Alexander Ku, Yinfei Yang, Burcu~Karagol Ayan, Ben
  Hutchinson, Wei Han, Zarana Parekh, Xin Li, Han Zhang, Jason Baldridge, and
  Yonghui Wu.
\newblock Scaling autoregressive models for content-rich text-to-image
  generation.
\newblock \emph{Transactions on Machine Learning Research}, 2022.
\newblock URL \url{https://arxiv.org/abs/2206.10789}.

\bibitem[Yu et~al.(2024)Yu, Lezama, Gundavarapu, Versari, Sohn, Minnen, Cheng,
  Birodkar, Gupta, Gu, Hauptmann, Gong, Yang, Essa, Ross, and
  Jiang]{yu2024magvit2}
Lijun Yu, Jos{\'e} Lezama, Nitesh~B. Gundavarapu, Luca Versari, Kihyuk Sohn,
  David Minnen, Yong Cheng, Vighnesh Birodkar, Agrim Gupta, Xiuye Gu,
  Alexander~G. Hauptmann, Boqing Gong, Ming-Hsuan Yang, Irfan Essa, David~A.
  Ross, and Lu~Jiang.
\newblock Language model beats diffusion -- tokenizer is key to visual
  generation.
\newblock In \emph{International Conference on Learning Representations
  (ICLR)}, 2024.
\newblock URL \url{https://arxiv.org/abs/2310.05737}.

\bibitem[Yu et~al.(2025)Yu, He, Deng, Shen, and Chen]{yu2024rar}
Qihang Yu, Ju~He, Xueqing Deng, Xiaohui Shen, and Liang-Chieh Chen.
\newblock Randomized autoregressive visual generation.
\newblock In \emph{IEEE/CVF International Conference on Computer Vision
  (ICCV)}, 2025.
\newblock URL \url{https://arxiv.org/abs/2411.00776}.

\bibitem[Yu et~al.(2020)Yu, Kumar, Gupta, Levine, Hausman, and
  Finn]{yu2020pcgrad}
Tianhe Yu, Saurabh Kumar, Abhishek Gupta, Sergey Levine, Karol Hausman, and
  Chelsea Finn.
\newblock Gradient surgery for multi-task learning.
\newblock In Hugo Larochelle, Marc'Aurelio Ranzato, Raia Hadsell,
  Maria{-}Florina Balcan, and Hsuan{-}Tien Lin (eds.), \emph{Advances in Neural
  Information Processing Systems 33: Annual Conference on Neural Information
  Processing Systems 2020, NeurIPS 2020, December 6-12, 2020, virtual}, 2020.
\newblock URL
  \url{https://proceedings.neurips.cc/paper/2020/hash/3fe78a8acf5fda99de95303940a2420c-Abstract.html}.

\bibitem[Zhai et~al.(2023)Zhai, Mustafa, Kolesnikov, and Beyer]{zhai2023siglip}
Xiaohua Zhai, Basil Mustafa, Alexander Kolesnikov, and Lucas Beyer.
\newblock Sigmoid loss for language image pre-training.
\newblock In \emph{IEEE/CVF International Conference on Computer Vision
  (ICCV)}, 2023.
\newblock URL \url{https://arxiv.org/abs/2303.15343}.

\bibitem[Zhang et~al.(2026)Zhang, Yu, Ma, Zhang, Pan, Yao, Xiao, Huang, Huang,
  and Zhao]{gcpo2026}
Guohui Zhang, Hu~Yu, Xiaoxiao Ma, JingHao Zhang, Yaning Pan, Mingde Yao, Jie
  Xiao, Linjiang Huang, Jie Huang, and Feng Zhao.
\newblock Group critical-token policy optimization for autoregressive image
  generation.
\newblock In \emph{International Conference on Learning Representations
  (ICLR)}, 2026.
\newblock URL \url{https://arxiv.org/abs/2509.22485}.

\bibitem[Zhang \& Tang(2026)Zhang and Tang]{taskmoe2025}
Jiaxing Zhang and Hao Tang.
\newblock Resolving task objective conflicts in unified model via task-aware
  mixture-of-experts.
\newblock In \emph{Proceedings of the 25th International Conference on
  Autonomous Agents and Multiagent Systems (AAMAS)}, 2026.
\newblock URL \url{https://arxiv.org/abs/2506.03591}.
\newblock Extended abstract.

\bibitem[Zhou et~al.(2025)Zhou, Yu, Babu, Tirumala, Yasunaga, Shamis, Kahn, Ma,
  Zettlemoyer, and Levy]{zhou2024transfusion}
Chunting Zhou, Lili Yu, Arun Babu, Kushal Tirumala, Michihiro Yasunaga, Leonid
  Shamis, Jacob Kahn, Xuezhe Ma, Luke Zettlemoyer, and Omer Levy.
\newblock Transfusion: Predict the next token and diffuse images with one
  multi-modal model.
\newblock In \emph{International Conference on Learning Representations
  (ICLR)}, 2025.
\newblock URL \url{https://arxiv.org/abs/2408.11039}.

\end{thebibliography}
\bibliographystyle{iclr2027_conference}

\appendix

\section{Extended related work}
\label{app:related}

\paragraph{Autoregressive visual generation and raster order.}
The premise that visual-token roles vary with position has a long history in
autoregressive image modeling. Image GPT \citep{chen2020imagegpt} predicts
pixels in raster order with a plain Transformer \citep{vaswani2017attention};
discrete tokenizers \citep{oord2017vqvae,esser2021vqgan,yu2024magvit2} made
the same recipe scalable to high-resolution text-to-image generation
\citep{ramesh2021dalle,yu2022parti,sun2024llamagen}, and masked-token
generation \citep{chang2022maskgit} decodes tokens in parallel over a few
iterations rather than in raster order. Two recent lines make position an
explicit design variable: next-scale prediction \citep{tian2024var} orders
tokens from coarse to fine so that early steps fix global structure, and
randomized autoregressive training \citep{yu2024rar} anneals from random
permutations to raster order and shows that the factorization order
materially affects sample quality. Rotary position encoding
\citep{su2024roformer} is the mechanism through which the backbones we study
bind a token to its raster index. Our map measures a consequence of this
positional structure that these works do not: how much the gradient from each
position conflicts with a second objective sharing the same parameters.

\paragraph{Unified models and unified tokenizers.}
Beyond the systems named in Section~\ref{sec:intro}, unified models differ
mainly in how the understanding pathway meets the visual-token stream.
SEED-LLaMA \citep{ge2024seedllama} and VILA-U \citep{wu2024vilau} tokenize
images into semantically aligned discrete codes so that a single next-token
objective serves both tasks. TokenFlow \citep{qu2024tokenflow} decouples
semantic and pixel-level codebooks behind shared indices, whereas UniTok
\citep{ma2025unitok} argues that reconstruction and semantic supervision do
not inherently conflict and enlarges the codebook instead. Show-o2
\citep{xie2025showo2} and BAGEL \citep{deng2025bagel} route generation through
continuous latents with flow matching; BAGEL uses a mixture-of-transformers
backbone \citep{liang2024mot} whose modality-specific parameters are a form of
the layer- and expert-level separation discussed in Section~\ref{sec:related},
and LMFusion \citep{shi2024lmfusion} freezes the language modules of a
pretrained LLM and trains parallel image modules, a separation motivated
explicitly by forgetting of text-only capabilities. Understanding-only models
trained by visual instruction tuning \citep{liu2023llava} are the reference
point for the understanding axis of all of these systems. Because the position
effect we measure is strongest when both objectives share the token sequence
(Section~\ref{sec:limitations}), we expect it to be largest for the SEED and
VILA-U family and smallest for designs that, like Janus-Pro, feed
understanding through a separate encoder.

\paragraph{Multi-task optimization and modality competition.}
Gradient-conflict remedies for generic multi-task learning include
Nash-bargaining aggregation of task gradients \citep{navon2022nashmtl},
alignment targets that pull the updates of related tasks together
\citep{wang2021gradvaccine}, and stochastic sign masking of inconsistent
gradient components \citep{chen2020graddrop}; the cosine between task
gradients that our map evaluates per position was proposed as an on-line
signal for weighting auxiliary losses by \citet{du2018gradsim}. Large-scale
re-evaluations \citep{kurin2022unitary,xin2022mto} find that many specialized
multi-task optimizers do not beat a well-tuned scalarization, which is one
reason we compare \pam{} against a matched-budget parameter-separation arm and
a random-position control rather than against generic gradient surgery alone.
The multimodal-learning literature documents that jointly trained multimodal
networks can underperform their unimodal counterparts because the modalities
learn at different rates \citep{wang2020multimodal}, and corrects the
imbalance by modulating each modality's gradient on the fly
\citep{peng2022ogm}; \pam{} applies the same modulation idea at the finer
resolution of token positions within one modality. Selecting which token
positions to train on has also been shown to matter in language-model
pretraining \citep{lin2024rho1}, which is consistent with our finding that
the position structure, not the act of modulating tokens, carries the \pam{}
effect.

\paragraph{Parameter-efficient adaptation and forgetting.}
All arms are trained with LoRA \citep{hu2022lora}, which is known to forget
less of the base model than full fine-tuning at the price of learning less
\citep{biderman2024lora}; the interaction of \pam{} with full fine-tuning is
listed as an open question in Section~\ref{sec:limitations}. Optimization uses
AdamW \citep{loshchilov2019adamw}, and the probe is implemented with
\texttt{torch.autograd} hooks \citep{paszke2019pytorch}.

\paragraph{Frequency-domain evaluation and statistics.}
The band-wise FID of Table~\ref{tab:freq} decomposes images with a Laplacian
pyramid \citep{burt1983laplacian}. Reporting the texture cost per frequency
band follows the observation that neural networks fit low-frequency structure
first \citep{rahaman2019spectral}, so an intervention that acts on early,
low-frequency positions is expected to trade against the high-frequency band.
Semantic decodability is measured with linear probes \citep{alain2016probes};
effect sizes are reported as partial $\eta^2$ \citep{cohen1988power} and
confidence intervals by paired bootstrap \citep{efron1993bootstrap}.

\section{Hyperparameters and implementation details}
\label{app:hyperparams}

Table~\ref{tab:hyper} lists the shared configuration. The probe runs inside
\texttt{torch.autograd} with forward hooks capturing module inputs and
tensor hooks on module outputs; per-bucket sums are accumulated as
rank-one products during the standard backward pass
(Eq.~\ref{eq:bucket}). The map is computed on a fixed diagnostic set of 8
batches (64 understanding + 64 generation samples each) under 3 dropout
seeds; 4 additional batches are held out for bucket selection and
threshold calibration (Section~\ref{sec:method}), so selection never sees
the reported map.

\begin{table}[h]
\centering
\small
\caption{Shared configuration.}
\label{tab:hyper}
\begin{tabular}{ll}
\toprule
Item & Value \\
\midrule
Backbones & Show-o-1.3B ($L{=}24$, $d{=}2048$, $T{=}256$); Janus-Pro-7B
($L{=}30$, $d{=}4096$, $T{=}576$) \\
Position buckets & $K=16$ contiguous raster buckets \\
LoRA & rank 16, $\alpha$ 32, dropout 0.05, $q,k,v,o$, all layers \\
Trainable parameters & 12.6M (Show-o), 31.5M (Janus-Pro) \\
Optimization & AdamW, lr $10^{-4}$, cosine, warmup 500, 20k steps, batch 128 \\
Task mixture & 1:1 understanding / generation \\
Diagnostic probe & 8 batches $\times$ (64+64), 3 dropout seeds \\
Probe cost & $1.21\times$ (Show-o) / $1.17\times$ (Janus-Pro) vs.\ standard backward \\
\pam{} & $m=6$, $\tau=0.13$, $\mathcal S=\{1,\dots,4,6,7\}$ (1-indexed),
$\hat u$ EMA refresh 2k steps \\
Down-weight variant & $w=0.25$ (+6.2 GenEval pts vs.\ +7.7 projection) \\
Training overhead & $+0.3\%$ wall-clock \\
Seeds & 3 per arm \\
\bottomrule
\end{tabular}
\end{table}

\section{ANOVA and bootstrap protocols}
\label{app:stats}

The ANOVA of Table~\ref{tab:anova} treats the map cells as observations of a
balanced design: each (layer, bucket) cell is observed once per module-group
(5 groups: attention $q,k,v,o$ and MLP) and per batch-seed replicate (24),
yielding $L\cdot K\cdot 5\cdot 24$ observations. Module and batch effects are
additive and balanced, so the error mean square is their pooled residual;
main-effect and interaction sums of squares follow the standard balanced
two-way decomposition, and partial $\eta^2$ is
$\mathrm{SS}_{\mathrm{effect}}/(\mathrm{SS}_{\mathrm{effect}}+\mathrm{SS}_{\mathrm{err}})$.
Degrees of freedom are $\mathrm{df}_{\mathrm{layer}}=L{-}1$,
$\mathrm{df}_{\mathrm{pos}}=15$,
$\mathrm{df}_{\mathrm{int}}=(L{-}1)\cdot 15$, and
$\mathrm{df}_{\mathrm{err}}=L\cdot K\cdot 5\cdot 24 - L\cdot K$.

Bootstrap CIs resample the GenEval prompt set and the MME item set
($10^4$ resamples) with paired assignment across arms, and the percentile
interval is reported. Seed variance is the standard deviation of the
arm metric across 3 seeds.

\section{Per-band position profiles}
\label{app:bands}

\begin{figure}[h]
  \centering
  \includegraphics[width=0.86\linewidth]{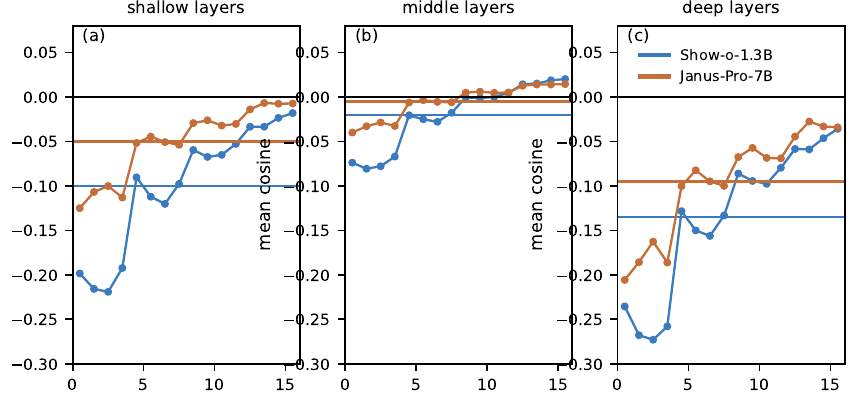}
  \caption{Position profiles of the interference map per layer band
  (shallow / middle / deep), both models; horizontal lines mark band
  means.}
  \label{fig:bands}
\end{figure}

Figure~\ref{fig:bands} plots the position profile of the conflict map
separately for the shallow, middle, and deep layer bands, with band means
$-0.10/-0.02/-0.135$ (Show-o) and $-0.05/-0.005/-0.095$ (Janus-Pro). The
middle band is nearly flat and slightly positive at late positions;
conflict is a joint function of depth band and position, which is what the
significant interaction in Table~\ref{tab:anova} quantifies
($\eta^2_{\mathrm{int}}=0.085/0.068$).

\section{Compute}
\label{app:compute}

\begin{figure}[h]
  \centering
  \includegraphics[width=0.86\linewidth]{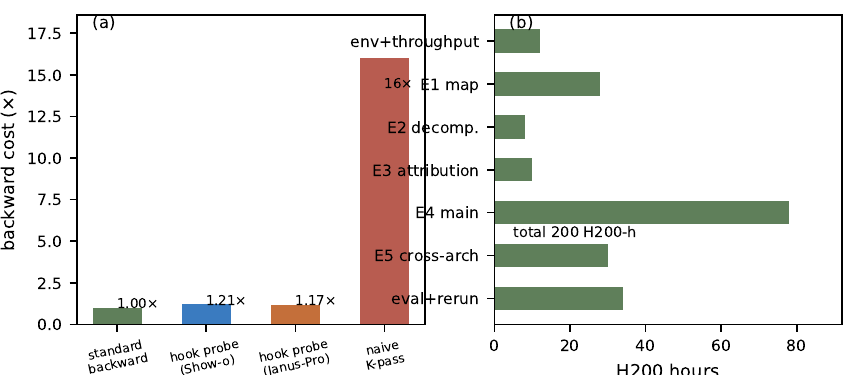}
  \caption{(a)~Backward cost of the probe vs.\ alternatives.
  (b)~H200-hour budget by stage; the total is 200 hours.}
  \label{fig:cost}
\end{figure}

Figure~\ref{fig:cost} breaks the total budget of 200 H200-hours: 12 for
environment bring-up and throughput measurement, 28 for the map (E1), 8 for
the norm--direction decomposition (E2, reusing E1 recordings), 10 for the
semanticity attribution (E3), 78 for the four-arm matched-budget training
(E4), 30 for the cross-architecture study (E5), and 34 for evaluation and
re-runs.

\section{Additional result figures}
\label{app:extra}

\begin{figure}[h]
  \centering
  \includegraphics[width=0.86\linewidth]{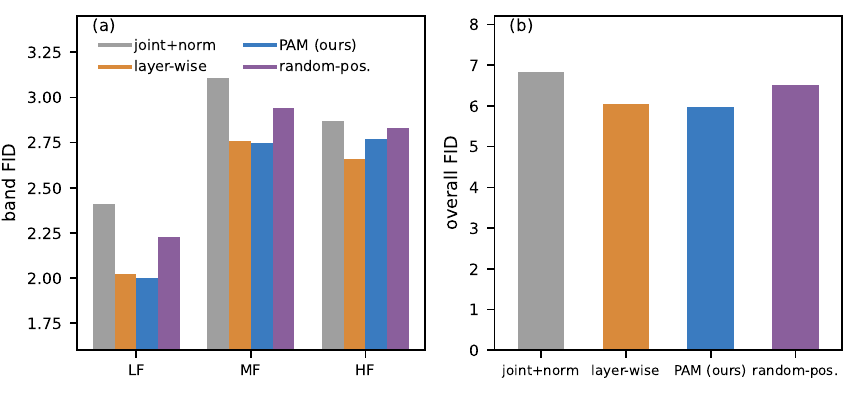}
  \caption{Frequency-band FID (a) and overall FID (b) on Show-o-1.3B for
  the four arms of Table~\ref{tab:freq}.}
  \label{fig:frequency}
\end{figure}

\begin{figure}[h]
  \centering
  \includegraphics[width=0.86\linewidth]{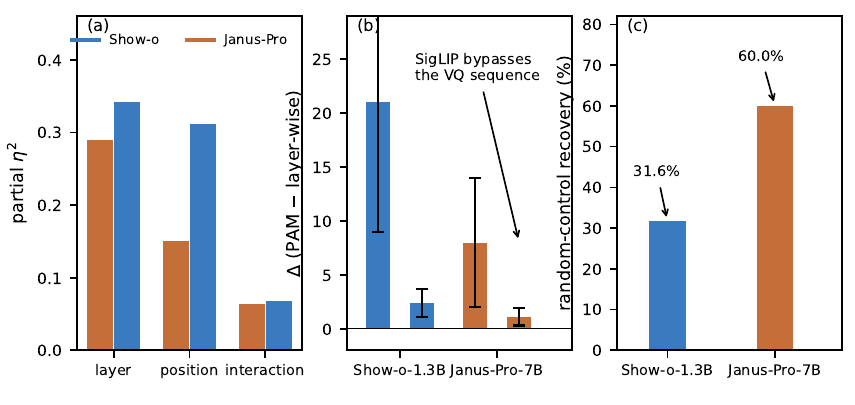}
  \caption{Cross-architecture comparison. (a)~ANOVA effect sizes by factor.
  (b)~Per-metric deltas of \pam{} over layer-wise separation with $95\%$
  CIs; on Janus-Pro the SigLIP understanding branch bypasses the visual
  token sequence and the effect halves. (c)~Random-control recovery of the
  \pam{} gain (MME).}
  \label{fig:cross}
\end{figure}

Figure~\ref{fig:frequency} shows the frequency-band FID decomposition of
Table~\ref{tab:freq}, including the high-frequency cost of \pam{}
($+4.1\%$ relative to layer-wise separation) against its low-frequency
gain ($-1.0\%$) and unchanged overall FID. Figure~\ref{fig:cross}
visualizes the cross-architecture comparison of Table~\ref{tab:cross}:
effect sizes by factor, per-metric deltas with CIs, and the
random-control recovery contrast ($31.6\%$ vs.\ $60.0\%$ on MME).

\end{document}